\documentclass[10pt,journal,compsoc]{IEEEtran}
\usepackage{algorithmic}
\usepackage{algorithm}
\usepackage{array}
\usepackage[caption=false,font=normalsize,labelfont=sf,textfont=sf]{subfig}
\usepackage{textcomp}
\usepackage{stfloats}
\usepackage{url}
\usepackage{verbatim}
\usepackage{graphicx}
\usepackage{multirow}
\usepackage{booktabs}
\usepackage{amssymb}
\usepackage{mathtools}
\usepackage{extpfeil}
\usepackage{diagbox}
\ifCLASSOPTIONcompsoc
  \usepackage[nocompress]{cite}
\else
  \usepackage{cite}
\fi
\ifCLASSINFOpdf
\else
\fi
\begin{document}
%
\title{STAU: A SpatioTemporal-Aware Unit for Video Prediction and Beyond}
\author{Zheng~Chang,
        Xinfeng~Zhang,~\IEEEmembership{Senior~Member,~IEEE,}
        Shanshe~Wang,~\IEEEmembership{Member,~IEEE,}
        Siwei~Ma,~\IEEEmembership{Senior~Member,~IEEE,}
        and~Wen~Gao,~\IEEEmembership{Fellow,~IEEE}
\thanks{Corresponding author: Siwei Ma (e-mail: swma@pku.edu.cn).}
\thanks{Z. Chang is with Institute of Computing Technology, Chinese Academy of Sciences,
Beijing 100190, China, with the University of Chinese Academy of Sciences, Beijing 100049, China,
and also with the National Engineering Laboratory for Video Technology, Peking University, Beijing 100871, China
(e-mail: changzheng18@mails.ucas.ac.cn).}
\thanks{X. Zhang is with the School of Computer Science and Technology,
University of Chinese Academy of Sciences, Beijing 100049, China
(email: xfzhang@ucas.ac.cn).}
\thanks{S. Wang and S. Ma are with the National Engineering Laboratory for Video Technology, Peking University, Beijing 100871, China, and also with the Information Technology R\&D Innovation Center of Peking University, Shaoxing 312000, China (email: sswang@pku.edu.cn; swma@pku.edu.cn).}
\thanks{W. Gao is with the National Engineering Laboratory for Video Technology, Peking University, Beijing 100871, China, and also with the University of Chinese Academy of Sciences, Beijing 100049, China (email: wgao@pku.edu.cn).}
}

\IEEEtitleabstractindextext{%
\begin{abstract}
Video prediction aims to predict future frames by modeling the complex spatiotemporal dynamics in videos. However, most of the existing methods only model the temporal information and the spatial information for videos in an independent manner but haven't fully explored the correlations between both terms. In this paper, we propose a SpatioTemporal-Aware Unit (STAU) for video prediction and beyond by exploring the significant spatiotemporal correlations in videos. On the one hand, the motion-aware attention weights are learned from the spatial states to help aggregate the temporal states in the temporal domain. On the other hand, the appearance-aware attention weights are learned from the temporal states to help aggregate the spatial states in the spatial domain. In this way, the temporal information and the spatial information can be greatly aware of each other in both domains, during which, the spatiotemporal receptive field can also be greatly broadened for more reliable spatiotemporal modeling. Experiments are not only conducted on traditional video prediction tasks but also other tasks beyond video prediction, including the early action recognition and object detection tasks. Experimental results show that our STAU can outperform other methods on all tasks in terms of performance and computation efficiency.
\end{abstract}

}

\maketitle

\IEEEdisplaynontitleabstractindextext

%
\IEEEpeerreviewmaketitle

\IEEEraisesectionheading{\section{Introduction}\label{sec:introduction}}

%
%
%
%
\IEEEPARstart{V}{ideo} prediction is playing an increasingly important role in representation learning area due to the great power in modeling spatiotemporal dynamics, and has great potential to be applied into a variety of multimedia applications, including video coding \cite{ma2019image}, autonomous driving \cite{hu2020probabilistic}, robot control \cite{finn2016unsupervised}, intelligent decision-making system \cite{oprea2020review}, etc. However, predicting the unknown future is a very challenging task, during which, the appearance information (spatial dynamic) and the motion information (temporal dynamic) in videos need to be jointly modeled. Motivated by the great achievement from the deep learning techniques in computer vision (CV) and natural language processing (NLP), deep learning-based methods have been employed for video prediction in recent years.

Videos can be treated as a special kind of sequential data, which are appropriate to be modeled by Recurrent Neural Networks (RNNs). Ranzato \emph{et al.} \cite{ranzato2014video} first utilized RNNs to model the spatiotemporal dynamics for videos in an unsupervised manner. To fix the gradient vanishing/exploding problems in RNNs, Long Short-Term Memories (LSTMs) \cite{hochreiter1997long,srivastava2015unsupervised,xingjian2015convolutional,finn2016unsupervised,oliu2018folded} and Gated Recurrent Units (GRUs) \cite{cho2014learning,ballas2016delving,shi2017deep,chang2021iprnn} were utilized to help stabilize the training process and capture long-term dependencies in videos. However, LSTMs and GRUs are mainly designed to capture the temporal dependencies, the spatial information in video frames is overlooked, indicating traditional RNN structure may not be powerful enough to handle the complex spatiotemporal dynamics in videos. To solve this problem, some works \cite{wang2017predrnn,wang2018predrnn++,wang2019memory,chang2021iprnn} attempted to redesign the structure of LSTMs and GRUs to help jointly model the spatial information (appearance details) and the temporal information (motion patterns) in videos.

However, the above works only focused on simple scenes, when it comes to more complex scenarios, the performance is usually far from satisfactory. In recent years, many works have been proposed to improve the model expressivity to more complex scenarios. Some works \cite{yu2019efficient,jin2020exploring} attempted to utilize the preserved visual details during the feature extraction to augment the visual quality of the predictions. \cite{guen2020disentangling,wu2021motionrnn} have explored to disentangle the physical dynamics (motion patterns) to help predict more satisfactory human actions. In addition, some works \cite{wang2018eidetic,chang2021astm,lin2020self,chang2021stae,lee2021video,chai2021cms} have begun to augment the long-term memorizing ability for LSTMs with the help of the attention mechanism, which can also help broaden the spatiotemporal receptive field of the predictive units. Furthermore, to handle the stochastic motion patterns in the real world, deep stochastic models have been proposed based on the Variational Auto-Encoders (VAEs) \cite{kingma2013auto,babaeizadeh2018stochastic,denton2018stochastic,franceschi2020stochastic,xu2020video,wu2021greedy}, which can help predict more reasonable motions from the latent space. Moreover, other works have improved the training scheme by employing the Generate Adversarial Networks (GANs) \cite{goodfellow2014generative,lee2018stochastic,kwon2019predicting,chen2020long,luc2020transformation} and other perceptual loss functions \cite{mathieu2016deep} for more naturalistic predictions.


Despite the significant improvements from the above methods, the temporal information and the spatial information are still utilized in an independent manner and the significant correlations between them are rarely discussed. Actually, in videos, the temporal information (motions) and the spatial information (appearance) are never independent but highly complementary to each other. An intelligent predictive unit should consider the important correlations between both terms. Take a simple scene as an example, where a person is running. In this case, on the one hand, the running motions account for the appearance variation, i.e. the temporal information impacts the spatial information. On the other hand, the changes in appearance can help human eyes perceive that the man is running, i.e. the spatial information impacts the temporal information. Although the output gate in LSTMs can help aggregate the temporal states and the spatial states, the correlations between both terms are still uncertain, restricting the model performance and expressivity in modeling reliable spatiotemporal dynamics for videos.
In this paper, we propose the spatiotemporal-aware unit to help prove that the spatiotemporal correlations in videos can greatly help improve the model performance.

Different from our previous work \cite{chang2021mau}, which only utilized the spatiotemporal correlations to augment the motions in the temporal domain, our STAU further utilizes the spatiotemporal correlations to augment the appearance in the spatial domain, as shown in Fig. \ref{fig:memory}. In particular, the STAU at time step $t$ cannot only perceive the spatiotemporal states in previous $\tau$ time steps (temporal domain), but also the spatiotemporal states in previous $\theta$ layers (spatial domain).
On the one hand, the motion-aware attention weights are learned from the spatial states from previous $\tau$ time steps, which are utilized to aggregate previous temporal states in the temporal domain.
On the other hand, the appearance-aware attention weights are learned from the temporal states in previous $\theta$ layers, which are further utilized to aggregate the spatial states in the spatial domain.
In this way, the temporal information and the spatial information can be greatly aware of each other in both the temporal and spatial domains.
The spatiotemporal correlations in both domains cannot only help build a motion-aware unit, but also an appearance-aware unit.
Moreover, STAU is more efficient with lower computation load and fewer parameters compared with existing predictive units and can be easily applied to other video prediction frameworks.

Overall, this journal paper mainly extends our previous work \cite{chang2021mau} in the following terms:
\begin{enumerate}
  \item We propose a spatiotemporal-aware unit (STAU) for video prediction and beyond, which is not only motion-aware but also appearance-aware.
  \item In the proposed STAU, not only the spatial information can supervise the temporal information in the temporal domain, but the temporal information can also supervise the spatial information in the spatial domain. In this way, the spatiotemporal correlations can be fully utilized in both the spatial and temporal domains.
  \item Experimental results show that the proposed STAU can outperform our previous work and existing methods in a variety of datasets. Besides, more ablation studies are conducted to evaluate the efficiency of the employed spatiotemporal correlations.
\end{enumerate}

The rest of the paper is organized as follows: Some related works are introduced in Section \ref{sec:related_work}. We elaborate on the model structure in Section \ref{sec:method}. The experimental results are summarized in Section \ref{sec:experiment} and Section \ref{sec:conclusion} concludes the whole paper.

\section{Related Work}\label{sec:related_work}
Deep learning techniques have shown their great capability in extracting efficient features for multimedia data and have been widely utilized for video prediction. Ranzato \emph{et al.} \cite{ranzato2014video} utilized Recurrent Neural Networks (RNNs) to model videos based on the language model. Srivastava \emph{et al.} \cite{srivastava2015unsupervised} improved RNNs with Long Short-Term Memories (LSTMs) to improve the model expressivity in capturing temporal dependency in videos, which is denoted as FC-LSTM. Shi \emph{et al.} \cite{xingjian2015convolutional} proposed ConvLSTM by replacing the fully-connected layers in FC-LSTM with convolutional layers to help improve the perceptions to visual data and save parameters. Similarly, Ballas \emph{et al.} \cite{ballas2016delving} also employed convolutional layers to the Gated Recurrent Units (GRUs) for video prediction. However, Wang \emph{et al.} \cite{wang2017predrnn} held the idea that the temporal information and the spatial information should be equally considered, and proposed a spatial module for ConvLSTMs (ST-LSTM) to help model the spatial representation for each frame. Then, they further proposed the Casual LSTM \cite{wang2018predrnn++} to help increase the model depth in the temporal domain and Gradient Highway Unit to alleviate the gradient propagation difficulties in deep predictive models. Guen \emph{et al.} proposed the PhyCell \cite{guen2020disentangling} to disentangle physical dynamics from unknown factors to predict more reliable motions. And Wu \emph{et al.} \cite{wu2021motionrnn} proposed the MotionGRU to independently model the transient variation and motion trend for more satisfactory predictions.

However, all the above methods only focus on predicting videos with simple spatiotemporal dynamics, and the performance in videos with complex scenarios is still far from satisfactory. To deal with this problem, on the one hand, some works attempted to preserve the spatiotemporal information outside the predictive memory for better predictions. Yu \emph{et al.} \cite{yu2019efficient} designed the conditionally reversible network (CrevNet) to preserve the spatiotemporal information during the feature extraction, and Jin \emph{et al.} \cite{jin2020exploring} preserved the spatiotemporal information in the frequency domain. On the other hand, since the attention mechanism can greatly help predictive units memorize long-term dependencies in natural language processing, some works have employed these methods for video prediction. Wang \emph{et al.} \cite{wang2018eidetic} proposed the recall gate to apply different levels of attention to different historical temporal states for long-term prediction, which was further improved by Chang \emph{et al.} in \cite{chang2021astm} by optimizing the computation process during the states aggregation. Lin \emph{et al.} \cite{lin2020self} proposed a self-attention memory (SAM) to memorize features with long-range dependencies in both temporal and spatial domains. And Chai \emph{et al.} \cite{chai2021cms} utilized attention mechanism to capture context correlations and multiscale spatiotemporal flows in video.

In spite of the noticeable improvements achieved by the above methods, videos are still predicted in a deterministic manner, which may be contradictory with the stochastic real-world scenarios. In recent years, Variational Auto-encoders (VAEs) \cite{kingma2013auto} have shown their great abilities in modeling the stochastic variables compared with the standard auto-encoders, motivated by which, stochastic video prediction methods are playing a more and more important role. Babaeizadeh \emph{et al.} \cite{babaeizadeh2018stochastic} proposed a stochastic variational video prediction (SV2P) method that predicts a different possible future for each sample of its latent variables. Denton \emph{et al.} \cite{denton2018stochastic} improved SV2P, which only utilized the fix prior $\mathcal{N}(0,I)$, with a learned prior for each time step (SVG). Xu \emph{et al.} \cite{xu2020video} further improved SVG by proposing a more reliable prior learning scheme with the help of the similar examples in the datasets. Franceschi \emph{et al.} \cite{franceschi2020stochastic} proposed a novel stochastic temporal model to untie frame synthesis and temporal dynamics via a residual update rule. Moreover, Wu \emph{et al.} \cite{wu2021greedy} proposed a Greedy Hierarchical Variational Auto-encoder, which is easy to train and can capture the multi-level stochasticity of future observations. Besides the stochastic methods, other methods \cite{lee2018stochastic,kwon2019predicting,chen2020long,luc2020transformation,mathieu2016deep} aim to improve the training process to generate more naturalistic predictions by utilizing Generative Neural Networks (GANs) and other perpetual loss functions.

In spite of the great achievements obtained by the above methods, the correlations between the temporal information and the spatial information have not been fully explored. In this paper, we propose the spatiotemporal-aware unit to deal with this problem.
\section{Method}\label{sec:method}
\begin{figure*}[t]
  \centering
  \includegraphics[width=\textwidth]{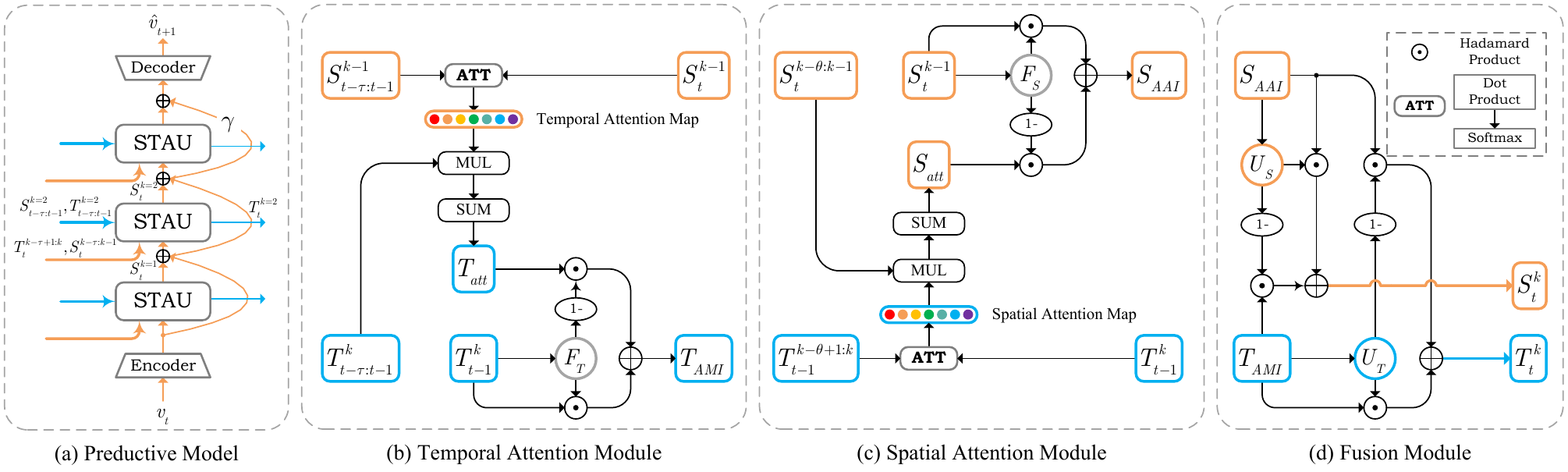}
  \caption{The structure of the spatiotemporal-aware unit. (a) The structure of the predictive model with multiple predictive units. (b) The temporal attention module, which is the main contribution of our previous work \cite{chang2021mau}. (c) The spatial attention module, which is our extended contribution based on our previous work \cite{chang2021mau}. (d) The fusion module, which aims to aggregate the spatial and the temporal information for the final predictions.}\label{fig:memory}
\end{figure*}
\subsection{Why exploring the spatiotemporal correlations?}\label{sec:why_cd}
Videos can be treated as an efficient mixture of temporal information and spatial information. On the one hand, the changing appearance (spatial information) over time contributes to the perception of the motions (temporal information). On the other hand, the changing motions (temporal information) contribute to the appearance (spatial information) of each frame. The spatiotemporal correlations may be the real source of videos, which are highly desired to be employed to model more reliable representations for videos.
However, such a strong correlation between the temporal information and the spatial information is difficult to be appropriately extracted via simply aggregating the temporal and spatial states, just as what previous works have done. The temporal information and the spatial information should first perceive each other, i.e. find the correlations. And then they should use the correlations to augment themselves.

Based on the above analysis, to fully utilize the spatiotemporal correlations to augment the spatiotemporal information for better predictions, two tasks need to be accomplished in both the temporal and spatial domains,
\begin{itemize}
  \item Extraction: how to extract the spatiotemporal correlations in both domains?
  \item Utilization: how to utilize the extracted spatiotemporal correlations to augment the temporal information and the spatial information in the corresponding domain?
\end{itemize}

In the following subsections, we will detailedly describe how the proposed spatiotemporal-aware unit (STAU) accomplishes the above tasks in both domains.
\subsection{The spatiotemporal-aware unit: STAU}\label{sec:stau}
In this section, we elaborate on how the proposed spatiotemporal-aware unit (STAU) extracts and utilizes the spatiotemporal correlations to augment the spatiotemporal information in the temporal domain and the spatial domain, respectively.

To extract and utilize the spatiotemporal correlations, the temporal information and the spatial information should know what they desire to get from each other. Previous works aim to accomplish this task via simply aggregating the spatial and temporal information using a convolutional neural network, which is inefficient and may lead to spatiotemporal information pollution. Herein, we design a more efficient way to solve this problem. In particular, we design two modules for STAU to help extract and utilize the spatiotemporal correlations in the temporal and spatial domains, respectively. In the temporal domain, we design the temporal attention module (TAM) to utilize the spatial information from different time steps to supervise the aggregation of the temporal information. In the spatial domain, we design the spatial attention module (SAM) to utilize the temporal information from different layers to supervise the aggregation of the spatial information. In this way, the temporal and spatial information can be greatly aware of each other and utilize the correlations to augment themselves, which is efficient and avoid the information pollution problem.

Like other recurrent predictive models, multiple STAUs are typically stacked to improve the model capability, as shown in Fig. \ref{fig:memory}(a), where $S_t^k$ denotes the spatial state at time step $t$ from layer $k$ and $T_t^k$ denotes the temporal state from time step $t$ in layer $k$. In particular, to extract the spatiotemporal correlations in both domains, for STAU at time step $t$ in layer $k$, spatiotemporal states from previous $\tau$ time steps and previous $\theta$ layers will be denoted as the inputs. In particular, for STAU in the bottom layer at time $t$, the spatial state is encoded from the video frame $v_t$, i.e. $S_t^{k=0}=Enc(v_t)$. Similarly, the predicted frame $\hat{v}_{t+1}$ at time step $t$ is decoded from the spatial state predicted by the STAU in the top layer $N$, i.e. $\hat{v}_{t+1}=Dec(S_t^{k=N})$.

In the following sections, we will elaborate on the detailed process to extract and utilize the spatiotemporal correlations in both the temporal and spatial domains, respectively.
\subsubsection{Extracting and utilizing spatiotemporal correlations in the temporal domain (Temporal Attention Module)}
As shown in Fig. \ref{fig:memory}(b), the temporal attention module aims to help the temporal states perceive the dynamics in the spatial states, which can help extract the spatiotemporal correlations in the temporal domain. The variations of the frame appearance over time help the human eyes have a perception to the motions in the temporal domain. To simulate this process, the contributions of the spatial states at previous time steps should be jointly considered while predicting the final temporal state at time step $t$. Especially, for current time step $t$, different spatial states from historical time steps can have different levels of contributions to current time step, which are defined as the motion-aware attention weights in our work. The motion-aware attention weights will be applied to the temporal state at corresponding time steps. In this way, the spatial information can supervise the information recombination process in the temporal domain. To implement this important process, the motion-aware attention weights between the spatial state at current time step and the ones at historical time steps are formulated as follows:
\begin{align}\label{equ:t_att}
  S^{\prime} &= W_s\ast S_t^{k-1},\nonumber\\
  p_i &= S_{t-i}^{k-1}\cdot S^{\prime},\nonumber\\
  \alpha_j &= \frac{e^{p_j}}{\sum_{i=1}^{\tau}e^{p_i}}.
\end{align}
where $\ast$ denotes the convolutional operator. The spatial state $S_t^{k-1}$ is first modulated to $S^{\prime}$ with a convolutional layer $W_s$. We use the dot product $\cdot$ to compute the correlation scores $p_i$. Then $p_i$ will be further normalized to the attention weights $\alpha_j$ using softmax operations.

To help the spatial information supervise the temporal information in the temporal domain, the attention scores $\alpha_j$ will be applied to the corresponding temporal states:
\begin{equation}
  T_{att} = \sum_{j=1}^\tau{\alpha_j\cdot T_{t-j}^k},
\end{equation}
where the temporal information $T$ can have a better knowledge of the spatial information $S$. $T_{att}$ denotes the temporal attention information (the temporal information which pays attentions to the spatial information), which can indicate the long-term motion trend, while $T_{t-j}^k$ only represents the short-term transient motion.

To efficiently aggregating the motion trend $T_{att}$ and the transient motion $T_{t-1}^k$ at time step $t$, the temporal fusion gate $F_T$ is initialized as follows,
\begin{align}\label{equ:u_fs}
  T^{\prime} &= W_t\ast T_{t-1}^k,\nonumber\\
  F_T &= \sigma(T^{\prime}),
\end{align}
where $T^{\prime}$ denotes the modulated temporal information, which will be also utilized in the spatial attention module. Then using the temporal fusion gate, the motion trend and the transient motion will be aggregated as follows,
\begin{equation}
  T_{AMI} = F_T\odot{T_{t-1}^k} + (1-F_T)\odot{T_{att}},
\end{equation}
where $\odot$ denotes the element-wise product and $T_{AMI}$ denotes the augmented motion information considering the spatiotemporal correlations. $F_T$ controls the information preserving percentage of the transient motion, and $1-F_T$ controls the information preserving percentage of the motion trend.
\subsubsection{Extracting and utilizing spatiotemporal correlations in the spatial domain (Spatial Attention Module)}
The spatial attention module concentrates on assisting the spatial information from different layers to perceive the temporal dynamics in the spatial domain, as shown in Fig. \ref{fig:memory}(c). The temporal information $T_{t-1}^k$ actually controls the appearance changing trend between the time step $t-1$ and $t$ in layer $k$. Since the spatial state $S_t$ directly indicates the appearance of frame $v_t$, we want to learn an attention map from the temporal states in the spatial domain, which can help the spatial states perceive the appearance changing trend. Using this prior knowledge, the appearance of each frame will change to a satisfactory direction.

\begin{table*}[b]
  \centering
  \setlength\tabcolsep{5pt}
  \caption{The experimental settings for different datasets on the video prediction task. \textbf{Train} and \textbf{Test} denote the temporal period of the inputs and predictions while training and testing. \textbf{Layers} denotes the number of the stacked predictive units. \textbf{Hidden} denotes the number of the channels of the hidden states. \textbf{DSL} and \textbf{USL} denote the downsampling layer and the upsampling layer, respectively, whose parameters are summarized in Table \ref{tab:parameter_setting}. $\gamma$ denotes the weights of the residual term in Equation \ref{equ:residual}.}
  \label{tab:experimental_setting}
  {\begin{tabular}{lcccccccc}
  \toprule
  \multicolumn{9}{c}{Experimental Settings}\\
  Dataset                                                                           &Resolution &Train &Test    &Layers &Encoder&Decoder&Hidden&$\gamma$\\
  \midrule
  Moving MNIST \cite{srivastava2015unsupervised}                                    &$64\times64\times1$    &$10\rightarrow10$ &$10\rightarrow10$&4&$2\times DSL$&$2\times USL$&64&0.0\\
  KITTI \cite{geiger2013vision}                                                     &$128\times160\times3$  &$10\rightarrow1$ &-&16&$3\times DSL$&$3\times USL$&64&1.0\\
  Caltech Pedestrian \cite{dollar2011pedestrian}                                    &$128\times160\times3$  &- &$10\rightarrow10$&16&$3\times DSL$&$3\times USL$&64&1.0\\
  TownCentreXVID \cite{benfold2011stable}                                           &$1088\times1920\times3$&$4\rightarrow1$ &$4\rightarrow4$&16&$4\times DSL$&$4\times USL$&64&1.0\\
  Something-Somethingv2 \cite{goyal2017something}                                   &$128\times128\times3$&$(5,10)\rightarrow(15,10)$ &$(5,10)\rightarrow(15,10)$&16&$3\times DSL$&$3\times USL$&64&1.0\\
  \bottomrule
  \end{tabular}}
\end{table*}
Similar with the temporal module, the attention weights in the spatial domain (appearance-aware attention weights) are formulated as follows,
\begin{align}
  q_i &= T_{t-1}^{k-i+1}\cdot T^{\prime},\nonumber\\
  \beta_j &= \frac{e^{q_j}}{\sum_{i=1}^{\theta}e^{q_i}},
\end{align}
where $T^{\prime}$ is defined in Equation \ref{equ:u_fs}. Using the computed attention score, the temporal state set can be aggregated as follows,
\begin{equation}
  S_{att} = \sum_{j=1}^\theta{\beta_j\cdot S_{t}^{k-j}}.
\end{equation}
$S_{att}$ is the spatial attention information, which can be treated as the global spatial information, having a knowledge of multi-level spatial information and the appearance changing rules from the temporal states.
To augment current spatial state (local spatial information) with the help of the global spatial information, a spatial fusion gate $F_S$ is defined as follows,
\begin{equation}
  F_S = \sigma(S^{\prime}),
\end{equation}
where $\sigma$ denotes the sigmoid function and $S^{\prime}$ is predefined in Equation \ref{equ:t_att}. Then using the spatial fusion gate, the local spatial information $S_t^{k-1}$ can be augmented by the global spatial information $S_{att}$,
\begin{align}
  S_{AAI} &= F_S\odot{S_t^{k-1}} + (1-F_S)\odot{S_{att}},
\end{align}
where $S_{AAI}$ denotes the augmented appearance information. $F_S$ controls the information preserving percentage of the local spatial information, and $1-F_S$ controls the information preserving percentage of the global spatial information.

\subsubsection{Fusion module}
With the help of the temporal attention module and the spatial attention module, the proposed model has greatly improved the ability to perceive the correlations between the temporal information and the spatial information. However, how to get the final predictions from the augmented information? Herein, we design the fusion module to further aggregate the augmented motion information $T_{AMI}$ and the augmented appearance information $S_{AAI}$. Different from the information updating rules in LSTMs, the information transitions in GRUs are more efficient, motivated by which, we design the information updating rules using two update gates,
\begin{align}
  U_T &= \sigma(W_{tu}\ast T_{AMI}),\nonumber\\
  U_S &= \sigma({W_{su}}\ast S_{AAI}),
\end{align}
where $U_T$ denotes the temporal update gate and $U_S$ denotes the spatial update gate. Using both gates, the aggregating process can be conducted as follows,
\begin{align}\label{equ:residual}
  T_t^k &= U_T\odot(W_{tt}\ast T_{AMI})+(1-U_T)\odot(W_{st}\ast S_{AAI}),\nonumber\\
  S_t^k &= U_S\odot(W_{ss}\ast S_{AAI})\nonumber\\&\quad+(1-U_S)\odot(W_{ts}\ast T_{AMI}) + \gamma\cdot S_t^{k-1},
\end{align}
where $U_T$ indicates the information preserving percentage of the temporal information and  $1-U_T$ indicates the information transition percentage from the spatial information to the temporal information. $U_S$ indicates the information preserving percentage from the spatial information and $1-U_S$ indicates the information transition percentage from the temporal information to the spatial information.
The residual-like term $\gamma\cdot S_t^{k-1}$ is utilized to stabilize the training process. The training details are further summarized in Alg. \ref{alg:training}.
\begin{algorithm}[t]
\caption{Training the proposed model for video prediction.}
\label{alg:training}
\textbf{Input}: $V:\{v_1,...,v_t,...,v_{T-1}\}$\\
\textbf{Parameter}: $W$\\
\textbf{Output}: $\hat{V}:\{\hat{v}_2,...,\hat{v}_t,...,\hat{v}_{T}\}$
\begin{algorithmic}[1] 
\REPEAT
\STATE $V:$ Random mini-batch from dataset
\STATE Let time step $t=1$, layer $k=1$, $\mathcal{L}_{MSE}=0.0$
\STATE Initializing $T_{t=0}^k, k={1\sim N}$ for STAUs
\WHILE{$t<T$}
\STATE $k=1$
\STATE $S_t^{k=0} = Enc(v_t)$
\WHILE{$k\leq N$}
\STATE $T_{input} = T_{\max(t-\tau,0):t-1}^k,T_{t-1}^{\max(k-\theta+1,0):k}$
\STATE $S_{input} = S_{\max(t-\tau+1,1):t}^{k-1}, S_t^{\max(k-\theta,0):k-1}$
\STATE $T_t^k, S_t^k= \text{\emph{STAU}}_k(T_{input} ,S_{input})$
\STATE $k=k+1$
\ENDWHILE
\STATE $\hat{v}_{t+1}=Dec(S_t^k)$
\STATE $\mathcal{L}_{MSE}+=\mathcal{L}_2(v_{t+1}, \hat{v}_{t+1})$
\STATE $t=t+1$
\ENDWHILE
\STATE $W\stackrel{+}{\longleftarrow}-\nabla_{W}\mathcal{L}_{MSE}$
\UNTIL{convergence}
\STATE \textbf{return} $\hat{V}:\{\hat{v}_2,...,\hat{v}_t,...,\hat{v}_{T}\}$
\end{algorithmic}
\end{algorithm}
\section{Experimental Results}\label{sec:experiment}
\subsection{Experimental Settings}

\begin{table*}[t]
  \centering
  \caption{The parameter settings for different layers. The encoder is built with the downsampling layers, which is utilized to extract deep features from video frames. The decoder is built with the upsampling layers, which is utilized to transform the features to the predicted frames. The hidden layer denotes the integrated convolutional layers in the predictive unit.}
  \label{tab:parameter_setting}
  {\begin{tabular}{lccccccc}
  \toprule
  \multicolumn{8}{c}{Parameter Settings}\\
  Layers                                 &operation &Kernel &Stride    &Padding &Features&Normalization&Nolinearity\\
  \midrule
  Downsampling Layer (Encoder)                                   &convolution&3&2&1&64&-&LeakeyRelu(0.2)\\
  Upsampling Layer (Decoder)                                    &deconvolution&3&2&1&64&-&LeakeyRelu(0.2)\\
  Hidden Layer (Unit)                                           &convolution&5&1&1&64&Layer Normalization&-\\
  \bottomrule
  \end{tabular}}
\end{table*}

In this section, extensive experiments are conducted to evaluate the performance of the proposed model compared with existing state-of-the-art methods. Besides traditional video prediction task, early action recognition task and object detection task are also conducted. Early action recognition task aims to recognize the actions after merely observing partial frames of a video, while object detection task aims to detect objects from the predictions using existing pre-trained algorithms, including Faster RCNNs \cite{ren2015faster}, Yolos \cite{redmon2016you}, etc. We utilized Yolov5s \cite{glenn_jocher_2021_4418161} pre-trained model in this paper. Both tasks can indirectly indicate the spatiotemporal modeling ability of the proposed predictive unit.

In particular, the video prediction task is conducted on 4 datasets, the Moving MNIST dataset \cite{srivastava2015unsupervised}, the Caltech Pedestrian dataset \cite{dollar2011pedestrian}, the KITTI dataset \cite{geiger2013vision}
and the TownCentreXVID dataset \cite{benfold2011stable}. The early action recognition task is conducted on Something-Somethingv2 dataset \cite{goyal2017something} and the object detection task is conducted on the predictions from the TownCentreXVID dataset. Moreover, video prediction models are optimized with the standard mean square error (MSE) loss function, averaged in all time steps, while early action recognition models are optimized with the cross-entropy (CE) loss function. All models are implemented with PyTorch and optimized using the Adam optimizer \cite{kingma2014adam} in a single Quadro RTX 4000 GPU (24GB). We set $\tau=5, \theta=5$ for all experiments. More detailed experimental settings for the above tasks are summarized in Table \ref{tab:experimental_setting}. In particular, for a fair comparison, we leave out the recalling scheme (introduced in our previous work \cite{chang2021mau}) in MAUs for all experiments.
\subsection{Video Prediction}
\begin{figure}[t]
  \centering
  \includegraphics[width=\columnwidth]{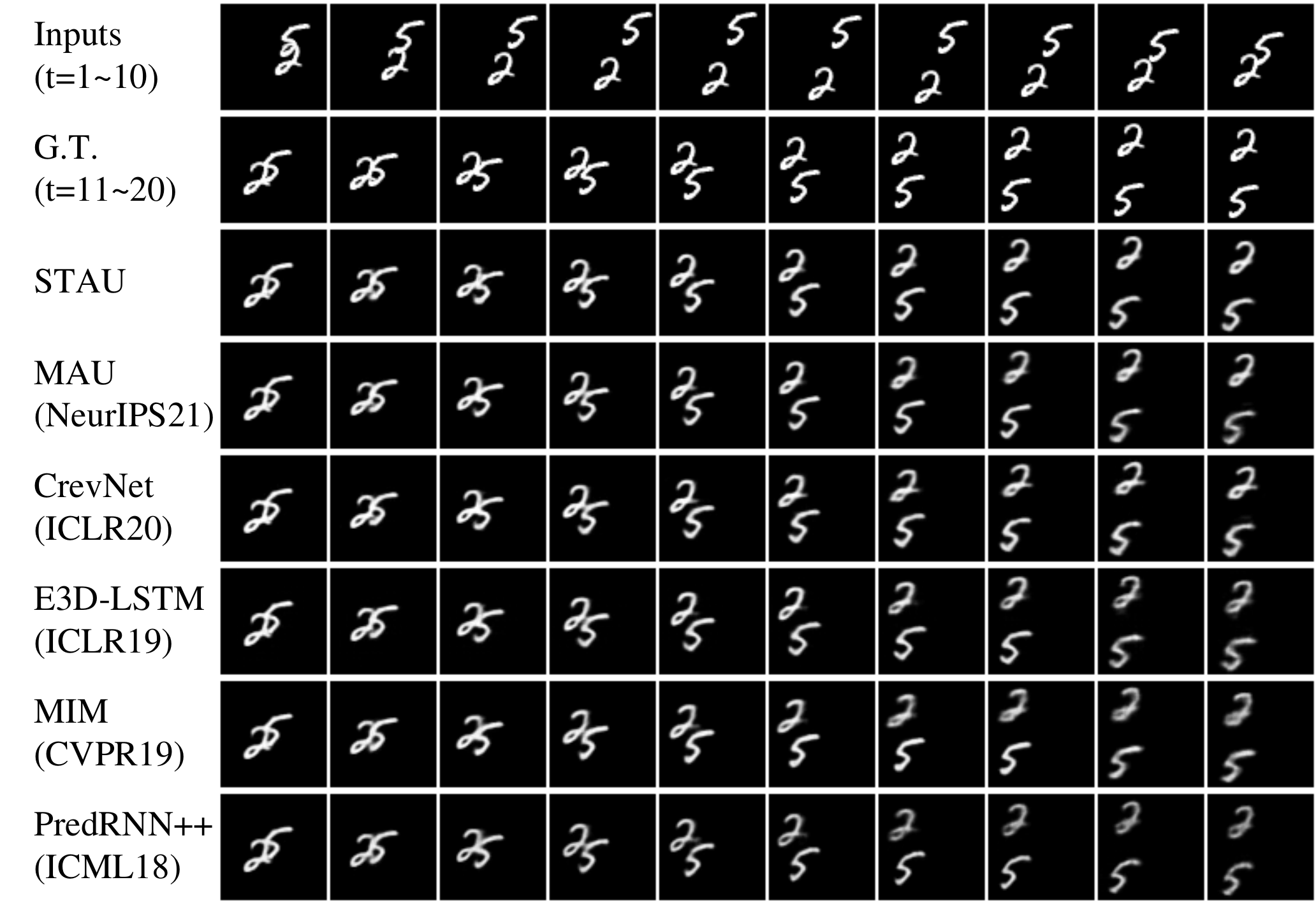}
  \caption{Predictions from different methods on the Moving MNIST dataset (10 frames $\longrightarrow $ 10 frames).}\label{fig:mnist}
\end{figure}
\begin{figure*}[t]
  \centering
  \includegraphics[width=\textwidth]{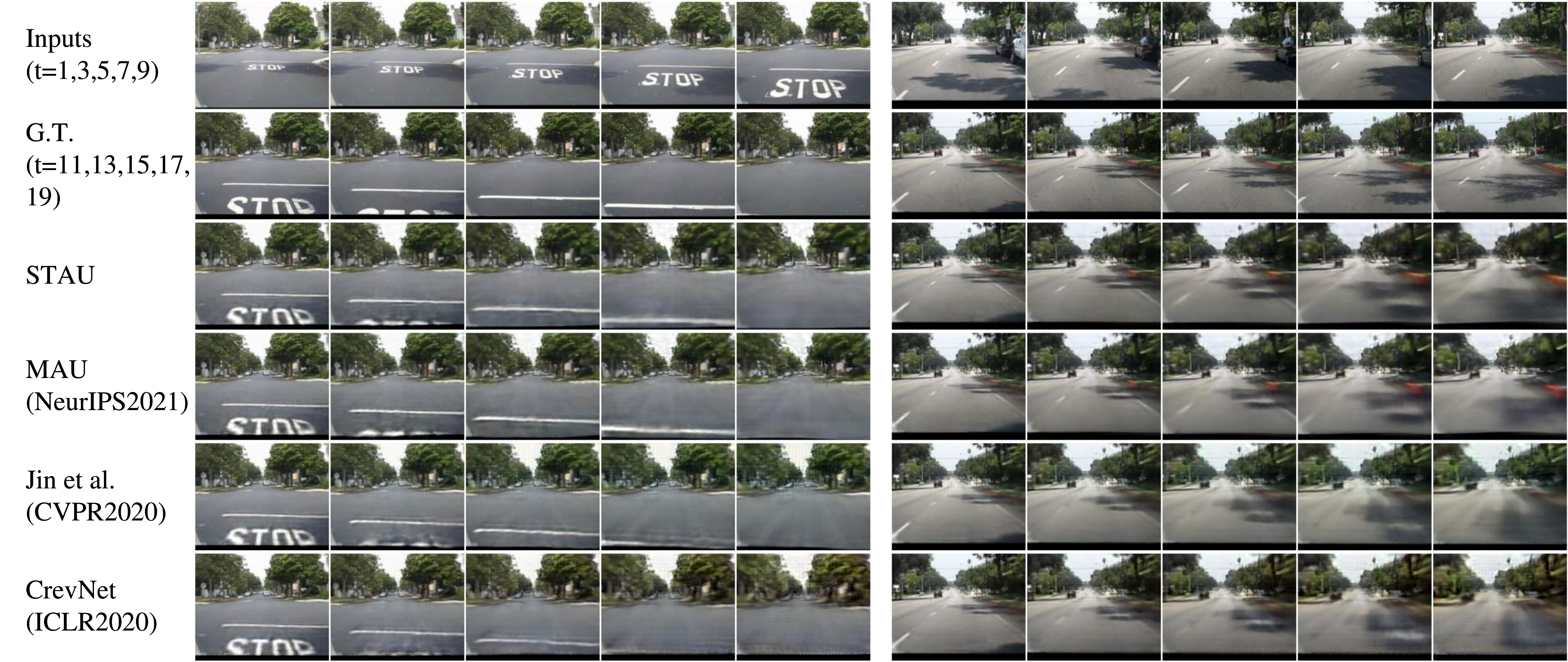}
  \caption{Predictions from different methods on the Caltech Pedestrian dataset (10 frames $\longrightarrow $ 10 frames).}\label{fig:caltech}
\end{figure*}
\begin{table}[t]
  \centering
  \caption{Quantitative results of different methods on the Moving MNIST dataset (10 frames $\rightarrow$ 10 frames). Lower MSE and higher SSIM scores indicate better visual quality. The results of the compared methods are reported in \cite{oprea2020review,chang2021mau}.}\label{tab:mnist}
  \begin{tabular}{lcc}
  \toprule
  \multicolumn{3}{c}{Moving MNIST}\\
  Method                                            &SSIM/frame $\uparrow$& MSE/frame $\downarrow$\\
  \midrule
  ConvLSTM (NeurIPS2015) \cite{xingjian2015convolutional}    & 0.707        & 103.3 \\
  FRNN (ECCV2018) \cite{oliu2018folded}                      & 0.819        & 68.4 \\
  VPN (ICML2017) \cite{kalchbrenner2017video}                & 0.870        & 70.0 \\
  PredRNN (NeurIPS2017) \cite{wang2017predrnn}               & 0.869        & 56.8 \\
  PredRNN++ (ICML2018) \cite{wang2018predrnn++}              & 0.898        & 46.5\\
  MIM (CVPR2019) \cite{wang2019memory}                       & 0.910        & 44.2 \\
  E3D-LSTM (ICLR2019) \cite{wang2018eidetic}                 & 0.910        & 41.3\\
  CrevNet (ICLR2020) \cite{yu2019efficient}                  & 0.928        & 38.5 \\
  MAU (NeurIPS2021) \cite{chang2021mau}                      & 0.931        & 29.5 \\
  \midrule
  STAU                                                       &\textbf{0.939} &\textbf{27.1}\\
  \bottomrule
  \end{tabular}
\end{table}
\begin{table*}[htb]
  \centering
  \caption{Quantitative results of different methods on the Caltech Pedestrian dataset. Lower MSE, LPIPS scores and higher SSIM, PSNR scores indicate better frame-level visual quality (10 frames $\rightarrow$ 1 frame). Lower FVD score indicates better sequence-level visual quality (10 frames $\rightarrow$ 10 frames). The results of the compared methods are reported in \cite{oprea2020review,chang2021mau}.}\label{tab:caltech}
  {\begin{tabular}{lccccc}
  \toprule
  \multicolumn{6}{c}{Caltech Pedestrian}\cr
  Method                     & MSE($10^{-3}$) $\downarrow$ & SSIM $\uparrow$ & PSNR $\uparrow$&LPIPS($10^{-2}$) $\downarrow$&FVD/10 frames $\downarrow$\\
  \midrule
  BeyondMSE (ICLR2016) \cite{mathieu2016deep}                                    & 3.42         & 0.847     &-      &-      &-   \\
  MCnet (ICLR2017) \cite{villegas2017decomposing}                                & 2.50         & 0.879     &-      &-      &-   \\
  CtrlGen (CVPR2018) \cite{hao2018controllable}                                  & -            & 0.900     &26.5   &-      &-  \\
  PredNet (ICLR2017) \cite{lotter2017deep}                                       & 2.42         & 0.905     &27.6   &9.89   &2860.8   \\
  ContextVP (ECCV2018) \cite{byeon2018contextvp}                                 & 1.94         & 0.921     &28.7   &9.53   &2451.6   \\
  E3D-LSTM (ICLR2019) \cite{wang2018eidetic}                                     & 2.12         & 0.914     &28.1   &10.02  &2311.2  \\
  Kwon \emph{et al.} (CVPR2019) \cite{kwon2019predicting}                        & 1.61         & 0.919     &29.2   &8.03   &1663.2   \\
  CrevNet (ICLR2020) \cite{yu2019efficient}                                      & 1.55         & 0.925     &29.3   &9.11   &1709.6  \\
  Jin \emph{et al.} (CVPR2020)\cite{jin2020exploring}                            & 1.59         & 0.927     &29.1   &8.99   &1441.1   \\
  MAU (NeurIPS2021) \cite{chang2021mau}                          & 1.34    &0.939 &29.4&8.51&1269.9\\
  \midrule
  STAU                          &\textbf{1.19}    &\textbf{0.944} &\textbf{29.9}&\textbf{7.54}&\textbf{950.6}\\
  \bottomrule
  \end{tabular}}
\end{table*}

\subsubsection{Moving MNIST}
Each sequence in the Moving MNIST dataset contains 20 successive frames with a resolution of $64\times64$, with 2 digits moving randomly. In particular, all models are trained and tested to predict the next 10 frames with the first 10 frames as the inputs. We use the Moving MNIST generating script\footnote{https://github.com/jhhuang96/ConvLSTM-PyTorch} to generate moving MNIST sequences from the standard MNIST training set \cite{lecun1998mnist}. Models are tested on the official Moving MNIST test set\footnote{http://www.cs.toronto.edu/~nitish/unsupervised\_video/}.

Fig. \ref{fig:mnist} shows the predicted examples from different methods, including our previous work and existing works. Compared with other methods, the proposed STAU has obtained predictions with the best visual qualities. In particular, compared with our previous work MAU \cite{chang2021mau}, the extended method STAU has obtained significantly better results in all time steps, especially the last three time steps, indicating better expressivity in spatiotemporal dynamics of videos. Moreover, some detailed quantitative results are summarized in Table \ref{tab:mnist}, where Mean Square Error (MSE) and Structural Similarity Index (SSIM) are utilized to indicate the visual quality of the predictions. As shown in Table \ref{tab:mnist}, the proposed STAU has achieved obviously better performance in all kinds of scores compare with our previous work and other existing methods.

\subsubsection{KITTI and Caltech Pedestrian}
KITTI \cite{geiger2013vision} and Caltech Pedestrian \cite{dollar2011pedestrian} are two real-world datasets. In particular, the KITTI dataset is collected for a variety of tasks, including stereo, optical flow, visual odometry, 3D object detection, and 3D tracking. The resolution of the raw data in KITTI is $375\times1242$. The Caltech Pedestrian dataset consists of approximately 10 hours of $640\times480$ 30Hz videos, taken from a vehicle driving through regular traffic in an urban environment. Following the experimental settings in \cite{lotter2016deep}, the proposed model is trained on the KITTI dataset and tested on the Caltech Pedestrian dataset. In particular, sequences of 20 frames have been sampled from ``City'', ``Residential'', and ``Road'' categories in the KITTI dataset and set06 $\sim$ set10 in the Caltech Pedestrian dataset. The frame rate of videos in the Caltech Pedestrian dataset is adjusted to the same as KITTI (10fps). All frames in both datasets have been center-cropped and resized to $128\times160$. Overall, a total of 32,373 sequences are for training and 7,725 sequences for testing.

Fig. \ref{fig:caltech} shows the qualitative results from different methods, where the proposed method has obtained more satisfactory traffic lines in sample 1 and more clear tree shadows in sample 2. In addition, quantitative results are summarized in Table \ref{tab:caltech}. Besides traditional MSE, PSNR, and SSIM scores, we further utilize the Learned Perceptual Image Patch Similarity (LPIPS) score \cite{zhang2018unreasonable} and the Fréchet Video Distance (FVD) score \cite{unterthiner2018towards} to indicate the perceptual quality and the sequence-level quality of the predictions, respectively.
Quantitative results show that the proposed method has achieved the best performance among all kinds of scores.
\begin{figure*}[t]
  \centering
  \includegraphics[width=\textwidth]{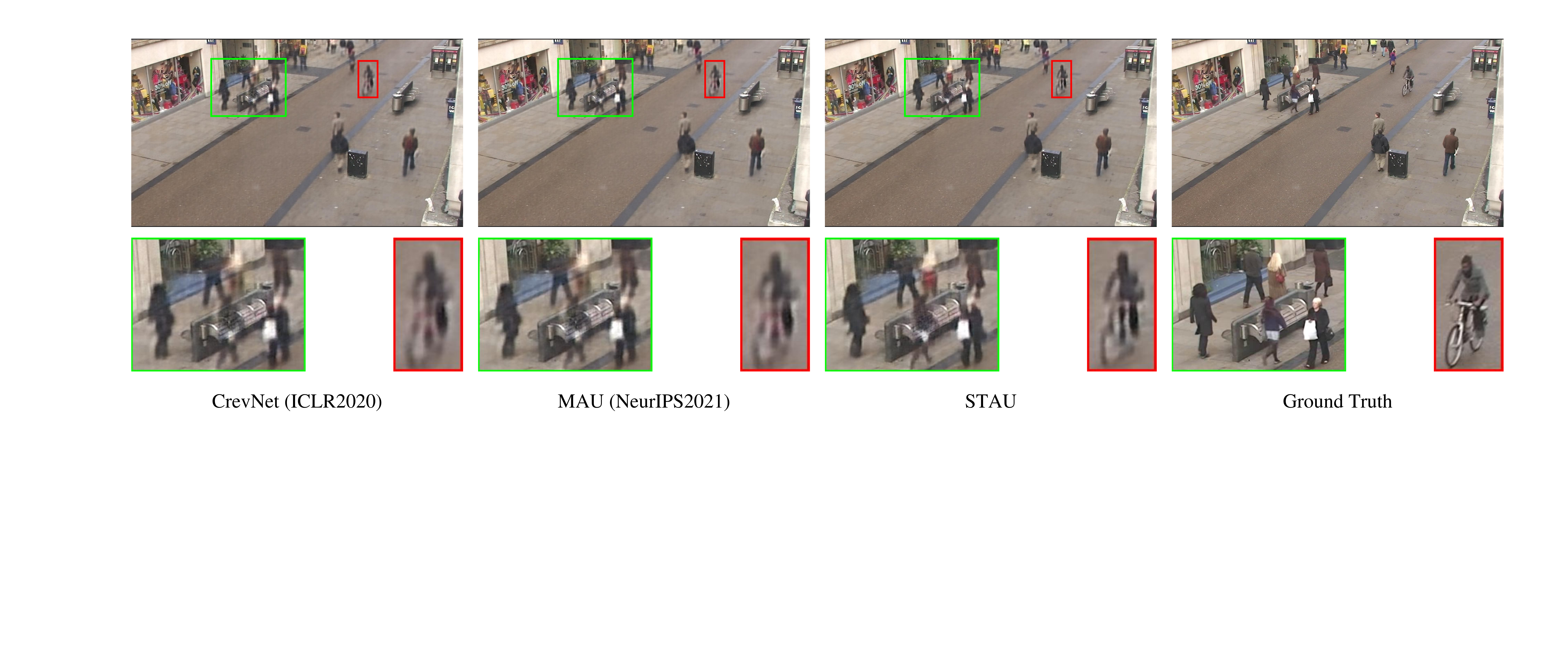}
  \caption{Predictions from different methods on the TownCentreXVID dataset (4 frames $\longrightarrow $ 1 frame).}\label{fig:town}
\end{figure*}
\begin{table*}[htb]
  \centering
  \caption{Quantitative results of different methods on the TownCentreXVID dataset (4 frames $ \rightarrow$ 4 frames). Higher SSIM and PSNR scores indicate better objective quality. Lower LPIPS score indicates better perceptual quality.}\label{tab:town}
   {\begin{tabular}{lcccccc}
    \toprule
    \multicolumn{7}{c}{TownCentreXVID}\cr
    Method
    &\multicolumn{3}{c}{$t=5$}&\multicolumn{3}{c}{$t=8$}\cr
    \cmidrule(lr){2-4}\cmidrule(lr){5-7}
    &PSNR $\uparrow$&SSIM $\uparrow$&LPIPS($10^{-2}$) $\downarrow$&PSNR $\uparrow$&SSIM $\uparrow$&LPIPS($10^{-2}$) $\downarrow$\cr
    \midrule
    ConvLSTM (NeurIPS2015) \cite{xingjian2015convolutional}             &27.22 &0.894  &39.90 &23.29 &0.876 &46.12     \cr
    PredRNN (NeurIPS2017) \cite{wang2017predrnn}                        &28.95 &0.921  &32.48 &23.82 &0.885 &37.85     \cr
    PredRNN++ (ICML2018) \cite{wang2018predrnn++}                       &29.50 &0.926  &30.59 &24.37 &0.894 &39.54     \cr
    E3D-LSTM (ICLR2019) \cite{wang2018eidetic}                          &29.70 &0.929  &29.47 &24.34 &0.901 &36.82     \cr
    CrevNet (ICLR2020) \cite{yu2019efficient}                           &30.12 &0.933  &27.87 &24.62 &0.910 &33.70     \cr
    MAU (NeurIPS2021) \cite{chang2021mau}                               &30.61 &0.937  &25.87  &25.52 &0.913 &32.42\cr
    \midrule
    STAU                                                                &\textbf{30.89}  &\textbf{0.943} &\textbf{21.95}    &\textbf{25.74} &\textbf{0.920} &\textbf{26.21}\\
    \bottomrule
    \end{tabular}}
\end{table*}
\begin{table*}[b]
  \centering
  \setlength\tabcolsep{3pt}
  \caption{Ablation study on the Moving MNIST dataset (10 frames $\rightarrow$ 10 frames). For fair comparison, the encoders and decoders are with the same structure for all models and all models are trained using Adam optimizer based on the MSE loss. }
  \label{tab:ablation_unit}
   { \begin{tabular}{lcccccc}
    \toprule
    Method&Backbone&MSE$\downarrow$&SSIM$\uparrow$&Parameters$\downarrow$/Memory&FLOPs$\downarrow$/Memory&Inference time$\downarrow$/800 samples\cr
    \midrule
    ConvLSTM (NeurIPS2015) \cite{xingjian2015convolutional}     &4$\times$ConvLSTMs         &102.1  &0.747  &0.98M  &\textbf{250.88M}&\textbf{16.47s}                \cr
    ST-LSTM (NeurIPS2017) \cite{wang2017predrnn}                &4$\times$ST-LSTMs          &54.5   &0.839  &1.57M  &401.92M&17.74s               \cr
    Casual-LSTM (ICML2018) \cite{wang2018predrnn++}             &4$\times$Casual-LSTMs      &46.3   &0.899  &1.80M  &460.80M&21.25s               \cr
    MIM (CVPR2019) \cite{wang2019memory}                        &4$\times$MIMs              &44.1   &0.910  &3.03M  &775.68M&45.13s               \cr
    E3D-LSTM (ICLR2019) \cite{wang2018eidetic}                  &4$\times$E3D-LSTMs         &40.1   &0.912  &4.70M  &1363.20M&57.21s                \cr
    RPM (ICLR2020) \cite{yu2019efficient}                       &4$\times$RPMs              &42.0   &0.922  &1.77M  &453.12M&18.01s                \cr
    MotionGRU (CVPR2021) \cite{wu2021motionrnn}                 &4$\times$MotionGRUs        &34.3   &0.928  &1.16M  &296.96M&17.58s                \cr
    MAU (NeurIPS2021) \cite{chang2021mau}                       &4$\times$MAUs              &29.5   &0.931  &0.78M  &265.37M&17.34s\cr
    STAU w/o TAM                                                &4$\times$STAUs             &29.9   &0.933  &0.78M  &265.37M&17.35s\\
    \midrule
    STAU                                                        &4$\times$STAUs             &\textbf{27.1} &\textbf{0.939}&\textbf{0.78M}&267.48M&19.58s\cr
    \bottomrule
    \end{tabular}}
\end{table*}

\subsubsection{TownCentreXVID}
To further evaluate the model performance in more complex real-world scenarios, TownCentreXVID dataset is utilized, containing video frames with more complex scenarios and higher resolution ($1080\times1920$). TownCentreXVID dataset contains 7,500 frames, where the first 4,000 frames are for training and the last 3,500 frames are for testing. In particular, to evaluate the model performance in high-resolution videos, all frames abandon the downsampling process in our experiments.

Fig. \ref{fig:town} shows the qualitative results from different methods, where the prediction from the proposed method can achieve the best visual quality. Table \ref{tab:town} shows that the proposed STAU has obtained the best objective (PSNR, SSIM) and perceptual (LPIPS) scores. All results indicate that the proposed method can achieve satisfactory performance in videos with relatively high resolution and has great potential to be applied into a variety of downstream tasks.

\subsubsection{Ablation study}
In this subsection, some ablation studies are conducted on the Moving MNIST dataset. Firstly, to make a fair comparison between the proposed STAU and other predictive units, we employ different units into the same predictive model, where all the compared predictive models contain the encoder and the decoder with the same structure and the only difference between them is the structure of the utilized predictive units, i.e. the backbones. Besides, all models consist of the same number of units and are optimized with Adam optimizer based on MSE loss function. In this way, the model performance can directly indicate the performance of the predictive unit. The results are summarized in shown in Table \ref{tab:ablation_unit}. On the one hand, the proposed STAU outperforms MAU (STAU w/o SAM) and STAU w/o TAM, indicating that the spatiotemporal correlations need to be employed in both the temporal and spatial domains and the extensions to our previous works are very necessary. On the other hand, STAU has also achieved the best performance with satisfactory parameters and relatively low computation load compared with existing predictive units.

\begin{figure}[t]
  \centering
  \includegraphics[width=\columnwidth]{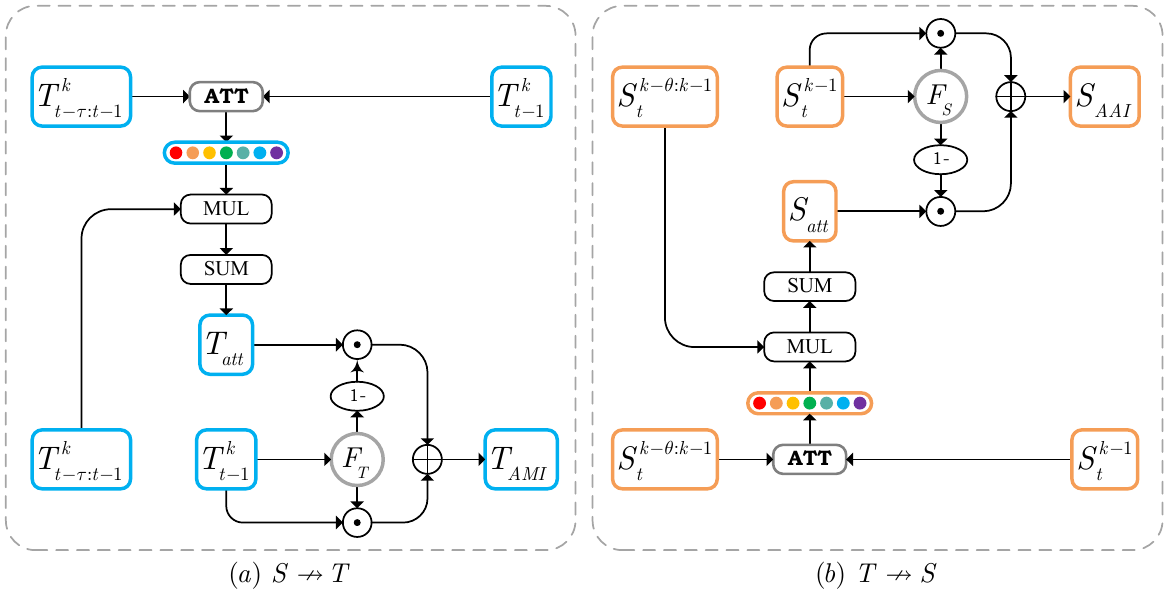}
  \caption{(a) The spatial information doesn't surprise the temporal information, where the spatiotemporal correlations are not utilized in the temporal domain. (b) The temporal information doesn't surprise spatial information, where the spatiotemporal correlations are not utilized in the spatial domain.}\label{fig:ablation_ST}
\end{figure}

Secondly, we evaluate the necessity of the proposed spatiotemporal correlations. We build another two modules, which don't utilize the spatiotemporal correlations in temporal and spatial domains, respectively, as shown in Fig. \ref{fig:ablation_ST}, where the temporal information and the spatial information are independently aggregated. We replace the SAM and the TAM in STAU with the above modules to evaluate the spatiotemporal correlations. All in all, the following 4 models are built for performance comparison.
\begin{enumerate}
  \item $S\leftrightarrow T$: Spatial information surprises temporal information (Fig. \ref{fig:memory}(b)), and temporal information also surprises spatial information (Fig. \ref{fig:memory}(c)).
  \item $S\rightarrow T, T\nrightarrow S$: Spatial information surprises temporal information (Fig. \ref{fig:memory}(b)), but temporal information doesn't surprise spatial information (Fig. \ref{fig:ablation_ST}(b)).
  \item $S\nrightarrow T,T\rightarrow S$: Spatial information doesn't surprise temporal information (Fig. \ref{fig:ablation_ST}(a)), but temporal information surprises spatial information (Fig. \ref{fig:memory}(c)).
  \item $S\nleftrightarrow T$: Spatial information doesn't surprise temporal information (Fig. \ref{fig:ablation_ST}(a)), and temporal information doesn't surprise spatial information (Fig. \ref{fig:ablation_ST}(b)).
\end{enumerate}
Results from different models are summarized in Table \ref{tab:ablation_ST}, where models utilizing spatiotemporal correlations outperform the ones without the spatiotemporal correlations. The results show that the spatiotemporal correlations play an important role in spatiotemporal predictive learning.

\begin{figure*}[t]
  \centering
  \includegraphics[width=\textwidth]{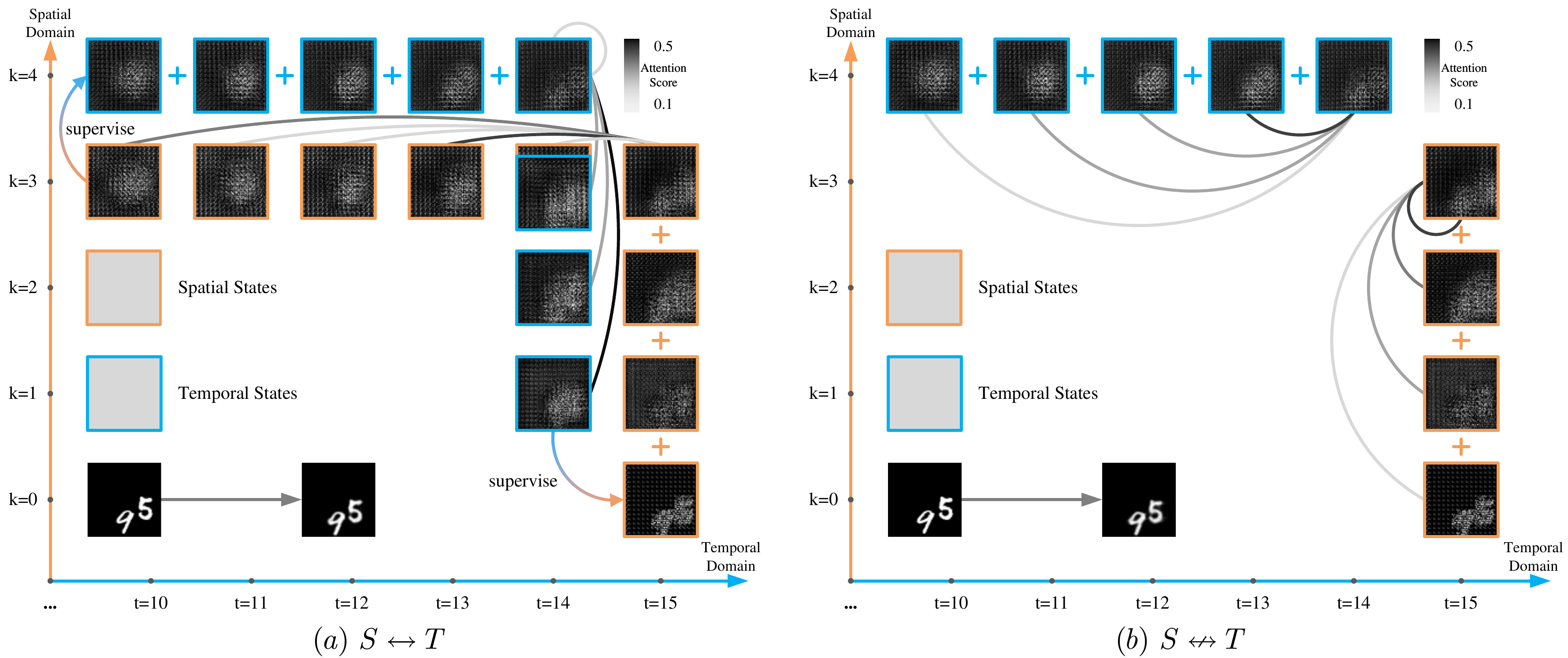}
  \caption{Visualized results of the attention scores and spatiotemporal states in both the temporal and spatial domains for STAU (a) and STAU without utilizing the spatiotemporal correlations (b) at time step 14 in layer 4. Feature maps with orange boundaries denote the spatial states and feature maps with blue boundaries denote the temporal states. All feature maps are normalized to $0\sim1$ and all states are placed following the scale in the temporal domain and the spatial domain. The colors of the attention flows follow the score maps in the top right corner (The darker the flow, the higher the attention score, the higher the importance of the state).}\label{fig:attention}
\end{figure*}

\begin{table}[htb]
  \centering
  \caption{Experimental results of models with different information interacting ways on the Moving MNIST dataset. $\rightarrow$ indicates supervising and $\nrightarrow$ indicates unsupervising.}
  \label{tab:ablation_ST}
   { \begin{tabular}{lcc}
    \toprule
    Method&MSE/frame $\downarrow$&SSIM/frame $\uparrow$\cr
    \midrule
    $S\nleftrightarrow T$                            &31.2 &0.927                 \cr
    $S\rightarrow T, T\nrightarrow S$               &29.5  &0.931           \cr
    $S\nrightarrow T,T\rightarrow S$                &29.6  &0.932            \cr
    $S\leftrightarrow T$                           &27.1  &0.939\\
    \bottomrule
    \end{tabular}}
\end{table}

To further evaluate the importance of the spatiotemporal correlations, we visualize the attention scores and the spatiotemporal states in both the temporal and spatial domains, as shown in Fig. \ref{fig:attention}. As shown in Fig. \ref{fig:attention}(a), in the temporal domain, the proposed STAU tends to pay more attentions to the temporal states in the longer history instead of the adjacent time step, indicating the predictive unit needs to observe a longer temporal period to predict more reliable motions. In the spatial domain, the proposed STAU pays the highest attention the low-level spatial states, and the lowest attention to the high-level spatial states, which indicates the predictive units aim to predict more reliable appearance information based on the low-level texture information and the multi-level (global) spatial information actually plays a more important role in frame prediction.
However, as shown in Fig. \ref{fig:attention}(b), the predictive unit without utilizing the spatiotmeporal correlations still only concentrates on the spatiotemporal states at the adjacent time step and the adjacent layer, because of which, the model performance are severely restricted.
All in all, the visualized results indicate the spatiotemporal correlations can help the predictive unit to pay higher attention to the more important spatiotemporal states instead of only the adjacent ones.

\begin{figure}[htb]
  \centering
  \includegraphics[width=\columnwidth]{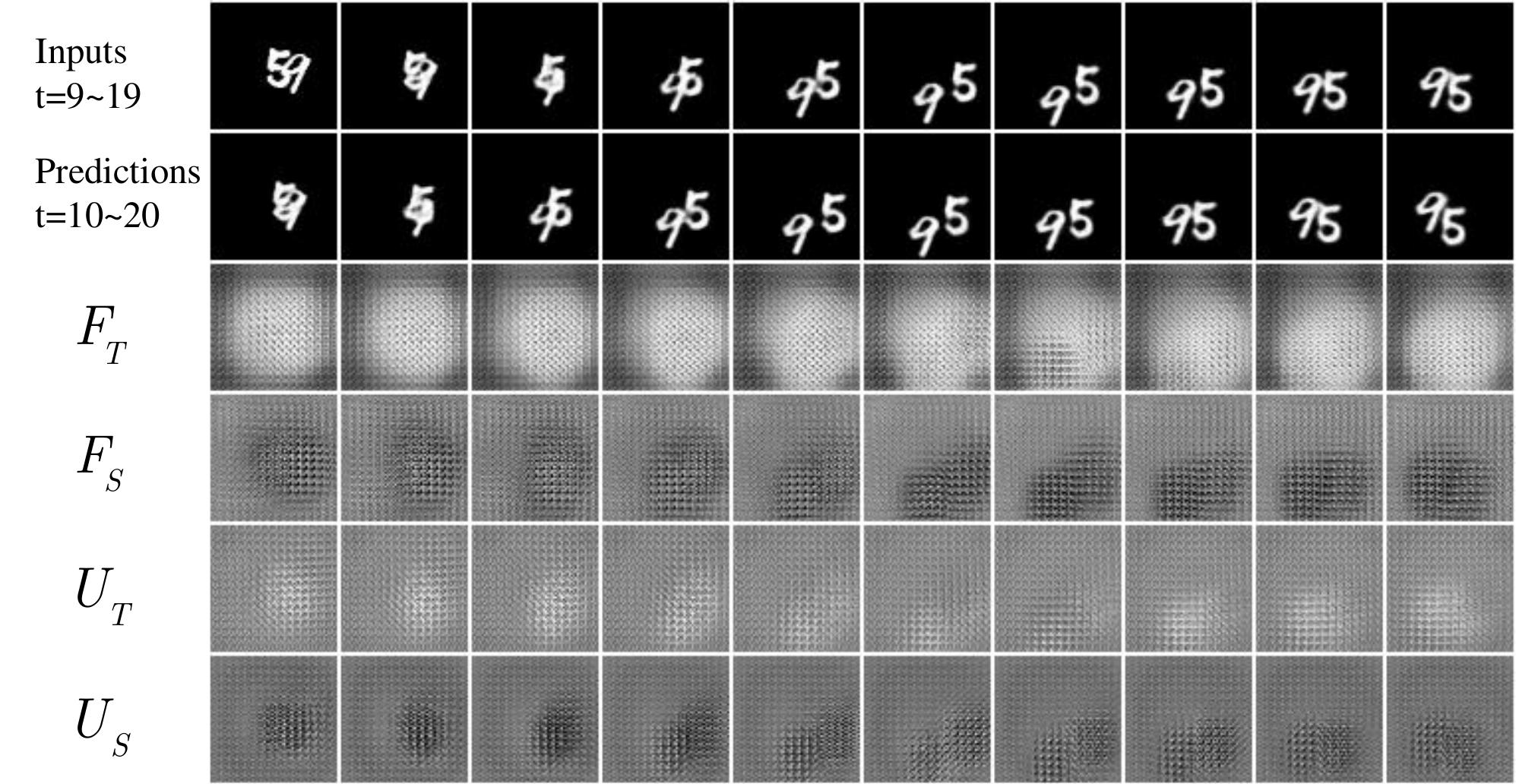}
  \caption{Visualized results of all the gates in the proposed STAU. The brighter the visualized gate is, the higher the weights of the gate are.}\label{fig:gates}
\end{figure}

To intuitively understand the information fusion process in the proposed STAU, we visualize the gates in the proposed predictive unit, as shown in Fig. \ref{fig:gates}. From the visualized results of the gates in STAU, the following conclusions can be obtained,
\begin{itemize}
  \item $F_T>F_S\longrightarrow $ In the temporal domain, the proposed STAU pays more attention to the transient motion $T_{t-1}^k$ than the motion trend $T_{att}$, indicating $T_{att}$ can help correct the transient motion for each time step but cannot decide the final motion direction. This result is also highly compatible with real world motion patterns. In the spatial domain, STAU concentrates on the global spatial information $S_{att}$ instead of the local spatial information $S_t^{k-1}$, which can also been proved in Fig. \ref{fig:attention}(a).
  \item $U_T>U_S\longrightarrow $ While predicting the final temporal state $T_t^k$, STAU tends to preserve more temporal information from the augmented motion information $T_{AMI}$. However, while predicting the final spatial state $S_t^k$, STAU is also more likely to utilize the temporal information from $T_{AMI}$. The above results indicate that $T_{AMI}$ is actually more important than $S_{AAI}$, which can also be proved by Table \ref{tab:ablation_unit}, where MAU (w $T_{AMI}$) outperforms STAU without TAM (w/o $T_{AMI}$).
\end{itemize}

\begin{figure}[t]
  \centering
  \includegraphics[width=\columnwidth]{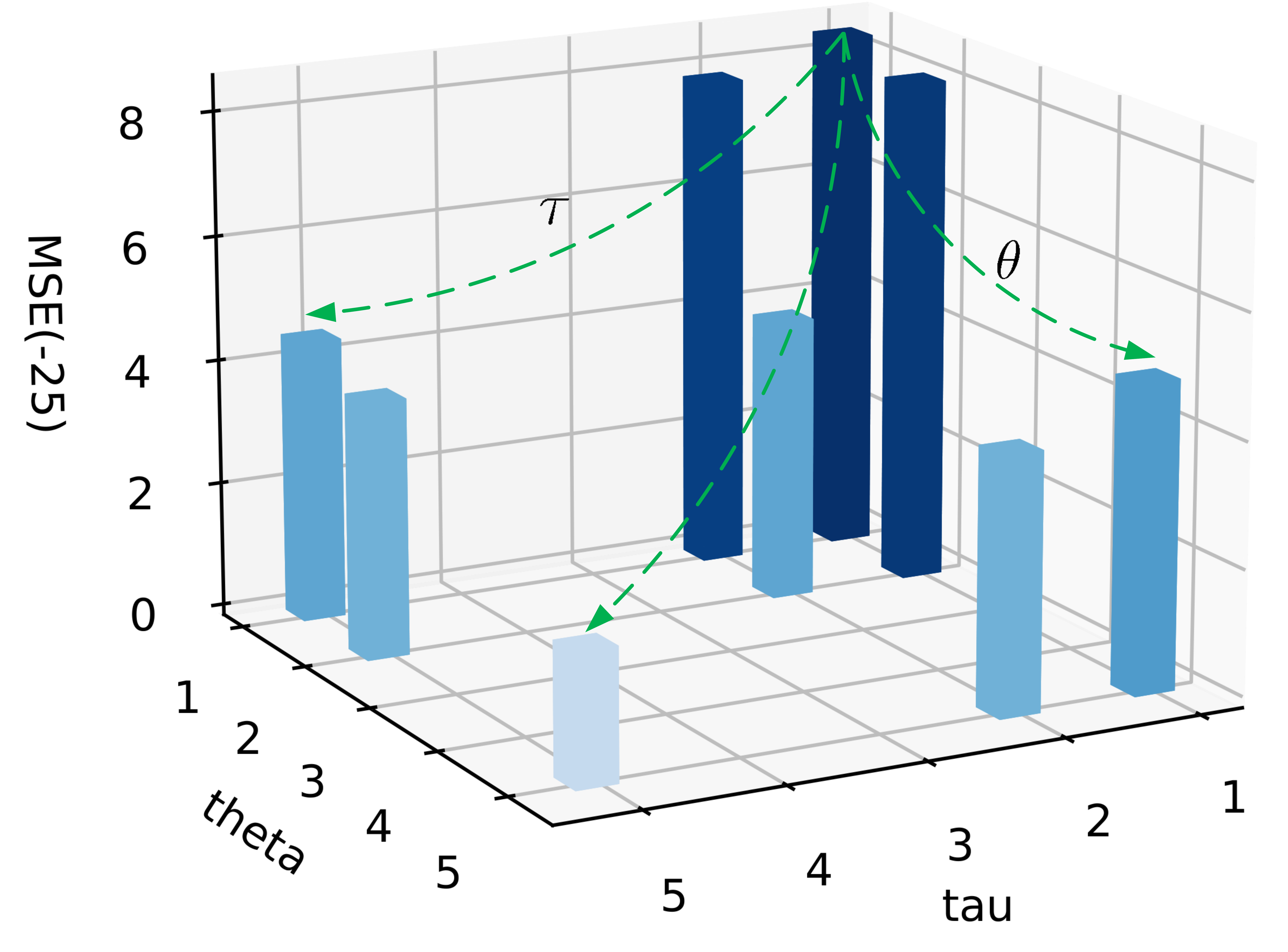}
  \caption{Experimental results of models with different size of spatiotemporal receptive fields.}\label{fig:receptive_field}
\end{figure}

Finally, since the proposed STAU needs to utilize multiple states in the temporal domain and the spatial domain, we want to explore the correlations between the spatiotemporal receptive field ($\tau,\theta$) and the model performance. As shown in Fig. \ref{fig:receptive_field}, model performance becomes better as the respective field $\tau,\theta$ increase, which indicates models can learn a more reliable representation for videos from a more broad receptive field in both the temporal and spatial domain. More detailed results are summarized in Table \ref{tab:receptive_field}.

\begin{table}[t]
  \centering
  \caption{Quantitative results (MSE scores) of different methods with different size of receptive fields on the Moving MNIST dataset.}\label{tab:receptive_field}
  \begin{tabular}{lcccc}
  \toprule
  \diagbox{$\theta$}{$\tau$}                                         &$\tau=1$  &$\tau=2$  &$\tau=5$\\
  \midrule
  $\theta=1$                                                      &33.4 &32.9  &29.5      \\
  $\theta=2$                                                      &33.1 &29.5  &29.1        \\
  $\theta=5$                                                      &29.9 &29.1   &27.1        \\
  \bottomrule
  \end{tabular}
\end{table}

\subsection{Beyond video prediction}
In this section, the proposed STAU is evaluated in some tasks beyond video prediction.
\subsubsection{Early Action Recognition: Something-Somethingv2}
The Something-SomethingV2 dataset is a large collection of labeled video clips that shows humans performing pre-defined basic actions with everyday objects. The whole dataset consists of 174 categories of videos. The training set contains 168,913 videos and the validation set consists of 24,777 videos. To evaluate the model performance in modeling spatiotemporal representations for videos, we conduct the early action recognition task on the Something-SomethingV2 dataset, where videos are categorized only based on part of the whole frames. Especially, in this paper, we utilize the front 25\% and 50 \% frames to categorize the whole video. Each video in the dataset is downsampled to 20 frames and each frame is resized to $128\times128$.

\begin{table*}[t]
  \centering
  \caption{The results of the early action recognition experiment of different methods on the Something-Something V2 dataset. Baseline method denotes that the classifier recognizes actions merely based on the front 25\% and 50\% frames without being extended by the video prediction model.}
  \label{tab:ss}
   { \begin{tabular}{lcccccc}
    \toprule
    \multicolumn{7}{c}{Something-SomethingV2}\cr
    \multirow{2}{*}{Method}&\multicolumn{3}{c}{Front 25\%}&\multicolumn{3}{c}{Front 50\%}\cr
    \cmidrule(lr){2-4}\cmidrule(lr){5-7}
    &PSNR $\uparrow$&top-1 $\uparrow$&top-5 $\uparrow$&PSNR $\uparrow$&top-1 $\uparrow$&top-5 $\uparrow$\cr
    \midrule
    Baseline                                                   &-           &2.10   &12.01  &-        &6.92   &14.29\cr
    ST-LSTM (NeurIPS2017) \cite{wang2017predrnn}               &14.87       &3.77   &14.17  &15.78    &8.91   &19.18\cr
    Casual-LSTM (ICML2018) \cite{wang2018predrnn++}            &15.34       &4.14   &14.67  &16.51    &9.57  &22.57\cr
    E3D-LSTM (ICLR2019) \cite{wang2018eidetic}                 &16.21       &4.76   &14.98  &17.05    &9.85  &24.24\cr
    RPM (ICLR2020) \cite{yu2019efficient}                      &16.53       &4.98   &15.07  &17.57    &10.01  &24.51\cr
    MotionGRU (CVPR2021) \cite{wu2021motionrnn}                &17.01       &5.11   &15.16  &17.86    &10.22  &27.65\cr
    MAU                                                        &17.36       &5.40   &19.00  &18.47    &10.60  &30.90\cr
    \midrule
    STAU                                                       &\textbf{17.50}   &\textbf{7.50}       &\textbf{21.80} &\textbf{18.48} &\textbf{12.90}&\textbf{32.30}             \\
    \bottomrule
    \end{tabular}}
\end{table*}
\begin{figure*}[htb]
  \centering
  \includegraphics[width=\textwidth]{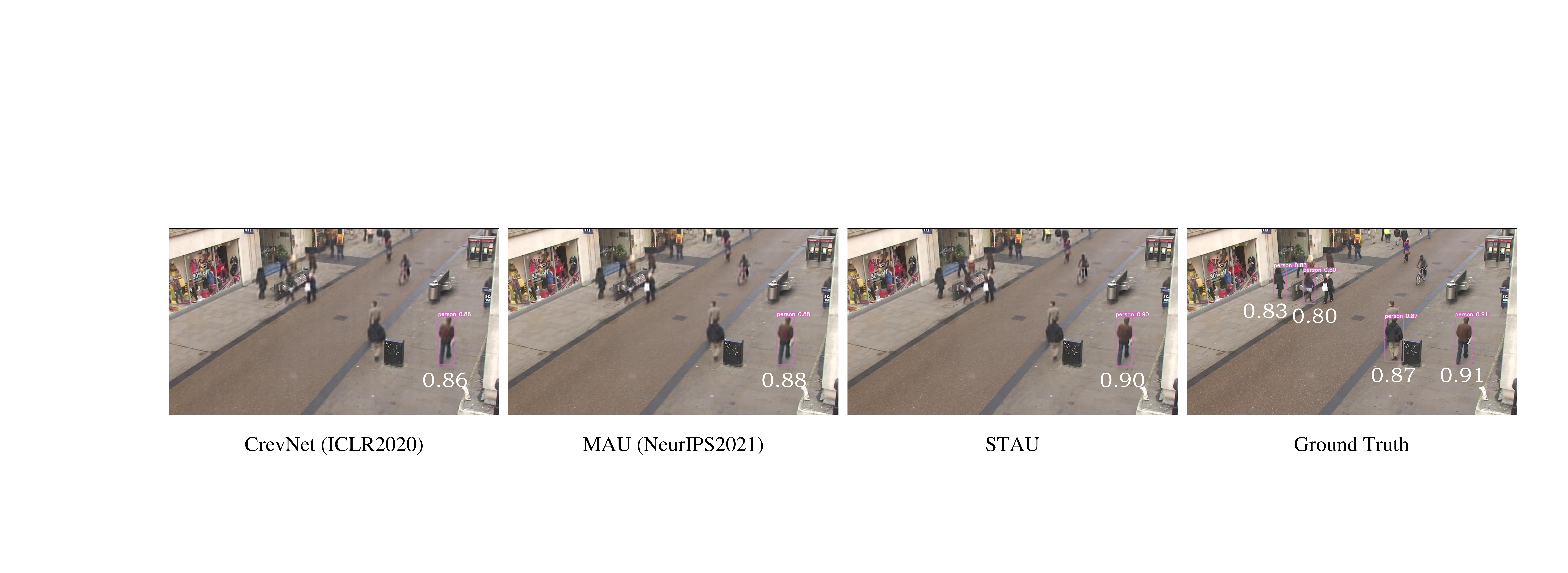}
  \caption{Object detection results on the predictions of different methods using Yolov5s pre-trained model \cite{glenn_jocher_2021_4418161}. The confidence threshold is set to 0.8.}\label{fig:town_dec}
\end{figure*}
In this task, models are first trained on the video prediction task to predict the next 15 and 10 frames using the first 5 and 10 frames. A total of 16 predictive memories, with 64 hidden channels, are stacked to form the predictive model. After training, 19 hidden states are concatenated to train the classifier, which is built on the basis of 3D convolutional layers. Then, the concatenated hidden states are transformed from $C\times T\times H\times W:64\times19\times16\times16$ to $174\times1$. Since reliable representations learned from the predictive units can lead to higher classification accuracy, the early action recognition task can greatly help indirectly evaluate the model performance in modeling spatiotemporal representations and the results are summarized in Table \ref{tab:ss}. On the one hand, models with predictive units obviously outperform the baseline model, indicating that video prediction models can help improve the performance in the early action task. On the other hand, the model with the proposed STAU has outperformed models with other predictive units in PSNR and top-1, top-5 scores, which indicates the proposed STAU can help learn a more reliable spatiotemporal representation for videos.

\subsubsection{Object Detection: TownCentreXVID}
In this section, we evaluated the proposed model on another important vision task, object detection. We conduct object detection task on the predictions from the TownCentreXVID dataset using the Yolov5s pre-trained model \cite{glenn_jocher_2021_4418161}. The confidence threshold is set to 0.8 and the detected confidence of the detected person can indicate the high-level visual quality of the predictions. The results are summarized in Fig. \ref{fig:town_dec}, where the confidence of the detected person in the prediction of the proposed STAU is the highest. The results of the object detection task show that the predictions of the proposed method not only benefit from the higher visual scores but also the higher performance in other tasks beyond video prediction, improving the practicability to other real-world applications.

\section{Conclusion}\label{sec:conclusion}
In this paper, we proposed a spatiotemporal-aware unit for video prediction and beyond. In particular, we designed the temporal attention module and the spatial attention module to model and utilized the spatiotemporal correlations in the temporal domain and the spatial domain, respectively. The state-to-state transitions in the proposed STAU are more efficient compared with existing LSTM-based and GRU-based predictive units. Extensive experimental results showed that the proposed STAU can outperform other state-of-the-art methods in related tasks by utilizing the spatiotemporal correlations. In addition, ablation studies also showed the extracted spatiotemporal correlations are necessary in improving the model expressivity in both domains.

\bibliographystyle{IEEEtran}
\bibliography{IEEEexample}
\begin{IEEEbiography}[{\includegraphics[width=1in,height=1.25in,clip,keepaspectratio]{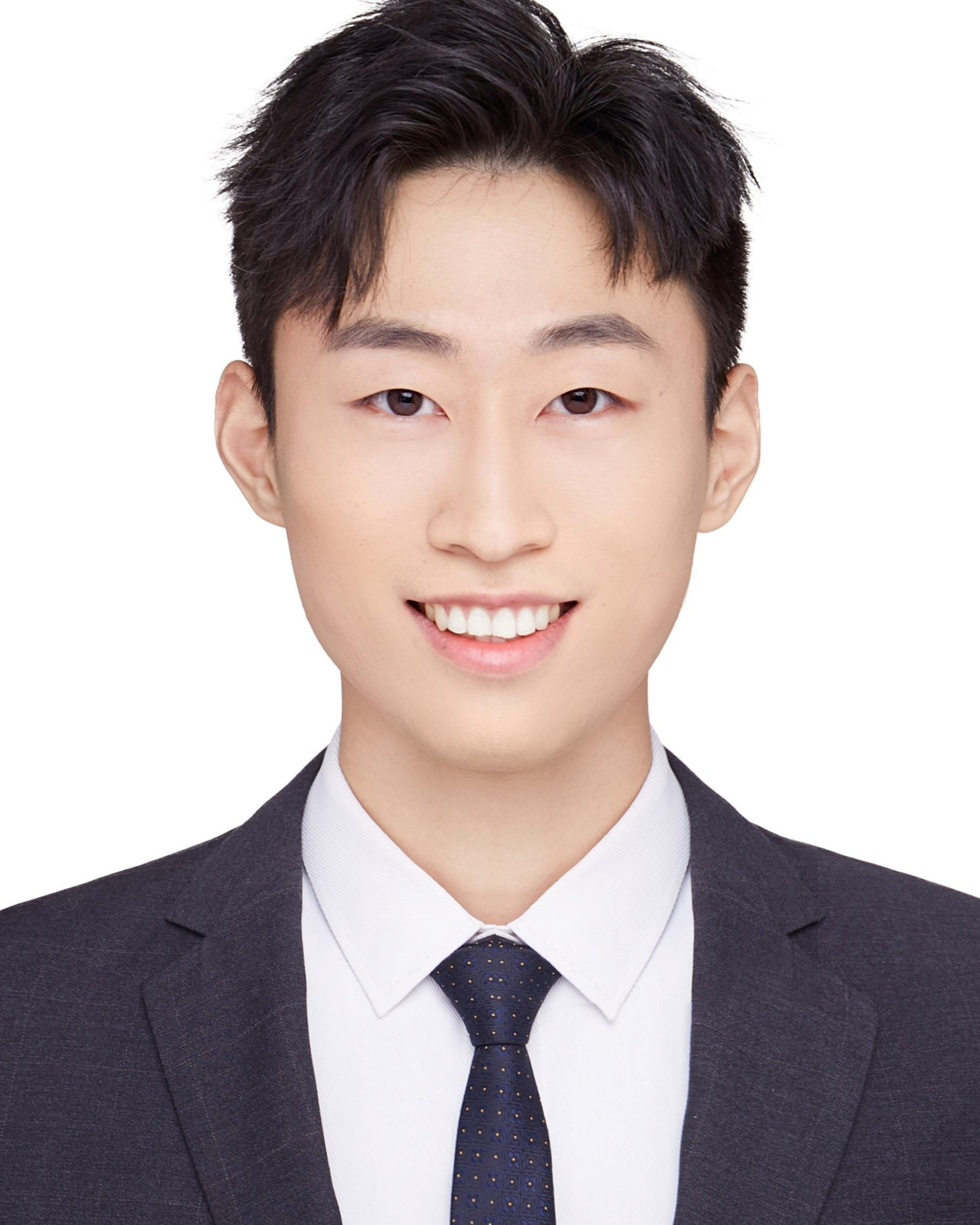}}]{Zheng Chang}
Zheng Chang received the B.E. degree in electronic information engineering from the University of Electronic Science and Technology of China, Chengdu, China, in 2018. He is currently pursuing the Ph.D. degree with the Institute of Computing Technology, Chinese Academy of Sciences, Beijing, China, and also with the University of Chinese Academy of Sciences, Beijing, China. His research interests include predictive learning and video understanding.
\end{IEEEbiography}
\begin{IEEEbiography}[{\includegraphics[width=1in,height=1.25in,clip,keepaspectratio]{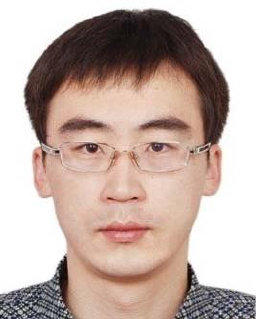}}]{Xinfeng Zhang}
Xinfeng Zhang (Senior Member, IEEE) received the B.S. degree in computer science from the Hebei University of Technology, Tianjin, China, in 2007, and the Ph.D. degree in computer science from the Institute of Computing Technology, Chinese Academy of Sciences, Beijing, China, in 2014.,From 2014 to 2017, he was a Research Fellow with the Rapid-Rich Object Search Laboratory, Nanyang Technological University, Singapore. From October 2017 to October 2018, he was a Postdoctoral Fellow with the School of Electrical Engineering System, University of Southern California, Los Angeles, CA, USA. From December 2018 to August 2019, he was a Research Fellow with the Department of Computer Science, City University of Hong Kong. He is currently an Assistant Professor with the School of Computer Science and Technology, University of Chinese Academy of Sciences. He has authored more than 150 refereed journal articles/conference papers. His research interests include video compression and processing, image/video quality assessment, and 3D point cloud processing. He received the Best Paper Award of IEEE Multimedia 2018, the Best Paper Award at the 2017 Pacific-Rim Conference on Multimedia (PCM), and the Best Student Paper Award in IEEE International Conference on Image Processing 2018.
\end{IEEEbiography}
\begin{IEEEbiography}[{\includegraphics[width=1in,height=1.25in,clip,keepaspectratio]{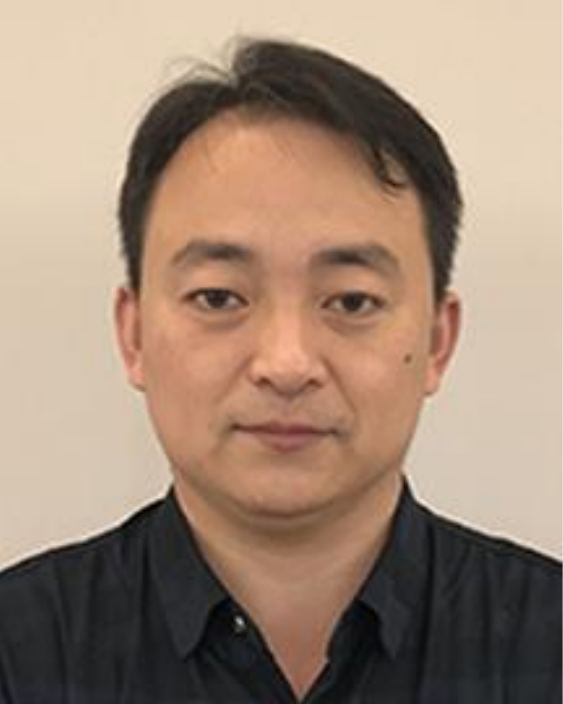}}]{Shanshe Wang}
Shanshe Wang (Member, IEEE) received the B.S. degree from the Department of Mathematics, Heilongjiang University, Harbin, China, in 2004, the M.S. degree in computer software and theory from Northeast Petroleum University, Daqing, China, in 2010, and the Ph.D. degree in computer science from the Harbin Institute of Technology. From 2016 to 2018, he held a postdoctoral position with Peking University, Beijing. He joined the School of Electronics Engineering and Computer Science, Institute of Digital Media, Peking University, where he is currently an Associate Researcher. His current research interests include video compression and image and video quality assessment.
\end{IEEEbiography}
\begin{IEEEbiography}[{\includegraphics[width=1in,height=1.25in,clip,keepaspectratio]{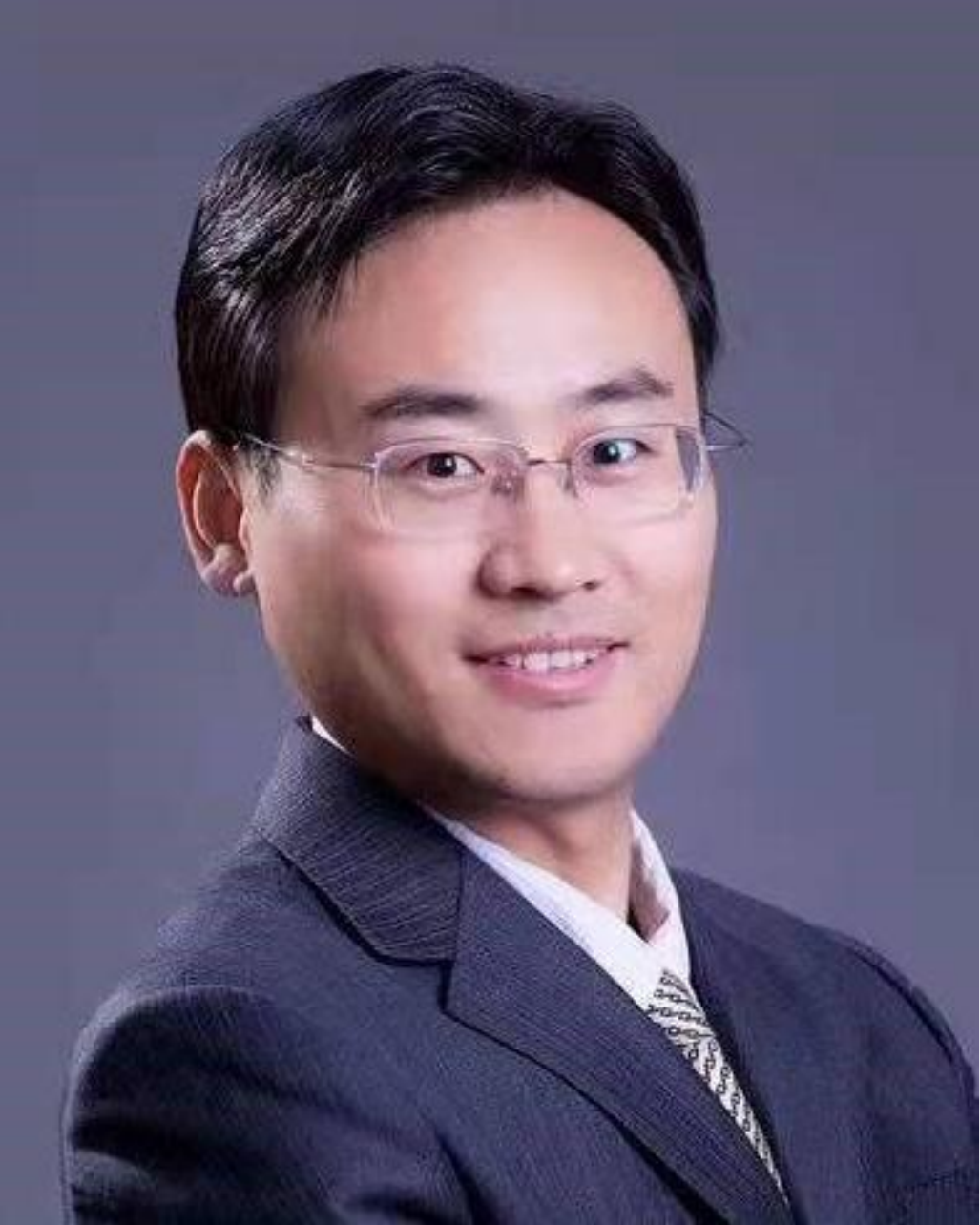}}]{Siwei Ma}
Siwei Ma (Senior Member, IEEE) received the B.S. degree from Shandong Normal University, Jinan, China, in 1999, and the Ph.D. degree in computer science from the Institute of Computing Technology, Chinese Academy of Sciences, Beijing, China, in 2005. He held a postdoctoral position with the University of Southern California, Los Angeles, CA, USA, from 2005 to 2007. He joined the School of Electronics Engineering and Computer Science, Institute of Digital Media, Peking University, Beijing, where he is currently a Professor. He has authored over 300 technical articles in refereed journals and proceedings in image and video coding, video processing, video streaming, and transmission. He served/serves as an Associate Editor for the IEEE Transactions on Circuits and Systems for Video Technology and the Journal of Visual Communication and Image Representation.
\end{IEEEbiography}
\begin{IEEEbiography}[{\includegraphics[width=1in,height=1.25in,clip,keepaspectratio]{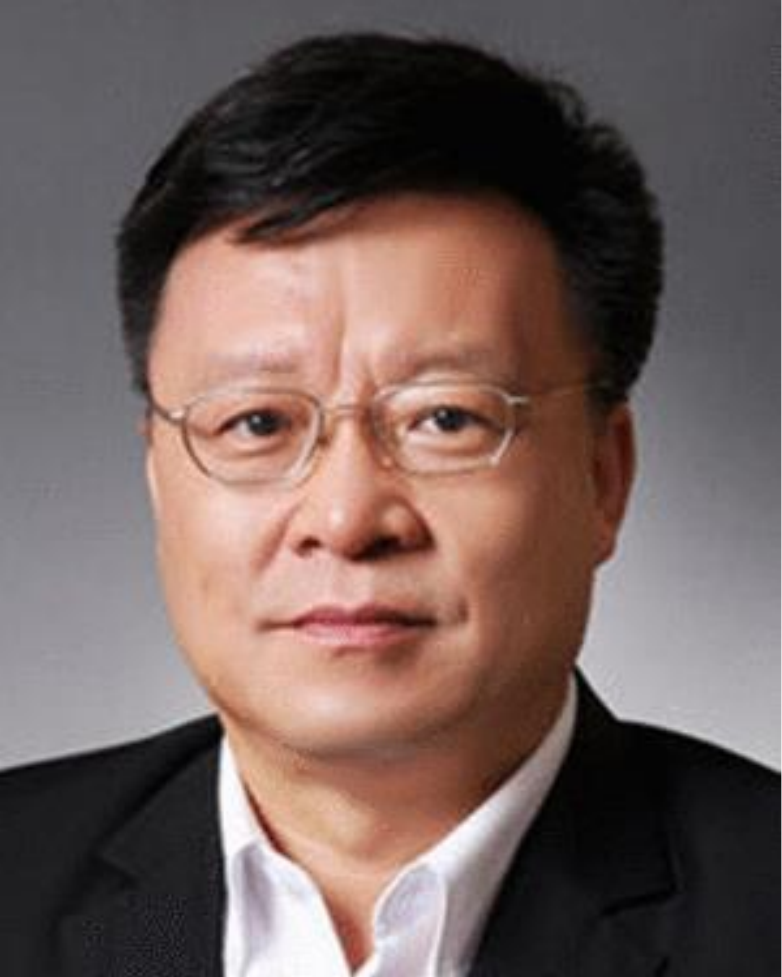}}]{Wen Gao}
Wen Gao (Fellow, IEEE) received the Ph.D. degree in electronics engineering from the University of Tokyo, Tokyo, Japan, in 1991. He is currently a Professor of computer science with the School of Electronic Engineering and Computer Science, Institute of Digital Media, Peking University, Beijing, China. Before joining Peking University, he was a Professor of computer science with the Harbin Institute of Technology, Harbin, China, from 1991 to 1995, and a Professor with the Institute of Computing Technology, Chinese Academy of Sciences, Beijing, China. He has authored or coauthored extensively, including five books and more than 600 technical articles in refereed journals and conference proceedings in the areas of image processing, video coding and communication, pattern recognition, multimedia information retrieval, multimodal interfaces, and bioinformatics. He is a Member of the China Engineering Academy. He is the Chair of a number of prestigious international conferences on multimedia and video signal processing, such as the IEEE International Conference on Multimedia and Expo and ACM Multimedia, and served on the Advisory and Technical Committees of numerous professional organizations. He served or serves on the Editorial Board of several journals, such as the IEEE Transactions on Circuits and Systems for Video Technology, IEEE Transactions on Multimedia, IEEE Transactions on Autonomous Mental Development, EURASIP Journal of Image Communications, and Journal of Visual Communication and Image Representation.
\end{IEEEbiography}
\end{document}